\documentclass[10pt,twocolumn,letterpaper]{article}

\usepackage{iccv}
\usepackage{times}
\usepackage{epsfig}
\usepackage{graphicx}
\usepackage{amsmath}
\usepackage{amssymb}
\usepackage{float}
\usepackage{multirow}

\usepackage{caption}
\usepackage{subcaption}
\usepackage{cite}
\usepackage{amsmath,amssymb,stmaryrd}   
\usepackage{multirow}

\usepackage[pagebackref=true,breaklinks=true,letterpaper=true,colorlinks,bookmarks=false]{hyperref}

\iccvfinalcopy

\usepackage{lipsum}

\newcommand\blfootnote[1]{%
  \begingroup
  \renewcommand\thefootnote{}\footnote{#1}%
  \addtocounter{footnote}{-1}%
  \endgroup
}

\ificcvfinal\fi

\begin{document}

\title{NEWTON: Neural View-Centric Mapping for On-the-Fly Large-Scale SLAM}

\author{Hidenobu Matsuki$^{1,2*}$, Keisuke Tateno$^{2}$, Michael Niemeyer$^{2}$, Federico Tombari$^{2,3}$\\
Imperial College London$^{1}$, Google$^{2}$, Tecnische Universität München$^{3}$
}

\maketitle
{\blfootnote{*This work was conducted during an internship at Google.}}

\ificcvfinal\thispagestyle{empty}\fi

\begin{abstract}

Neural field-based 3D representations have recently been adopted in many areas including SLAM systems.
Current neural SLAM or online mapping systems lead to impressive results in the presence of simple captures, but they rely on a world-centric map representation as only a single neural field model is used.
To define such a world-centric representation, accurate and static prior information about the scene, such as its boundaries and initial camera poses, are required. However, in real-time and on-the-fly scene capture applications, this prior knowledge cannot be assumed as fixed or static, since it dynamically changes and it is subject to significant updates based on run-time observations. Particularly in the context of large-scale mapping, significant camera pose drift is inevitable, necessitating the correction via loop closure. 
To overcome this limitation, we propose \textbf{NEWTON}, a view-centric mapping method that dynamically constructs neural fields based on run-time observation. In contrast to prior works, our method enables camera pose updates using loop closures and scene boundary updates by representing the scene with multiple neural fields, where each is defined in a local coordinate system of a selected keyframe. 
The experimental results demonstrate the superior performance of our method over existing world-centric neural field-based SLAM systems, in particular for large-scale scenes subject to camera pose updates.

\end{abstract}

\section{Introduction}

Real-time scene capture is an essential technology for embodied device scene understanding and interaction, and the choice of the underlying scene representation greatly influences the performance and efficiency of the mapping.

\begin{figure}[h]
        \centering
		\includegraphics[width=\linewidth]{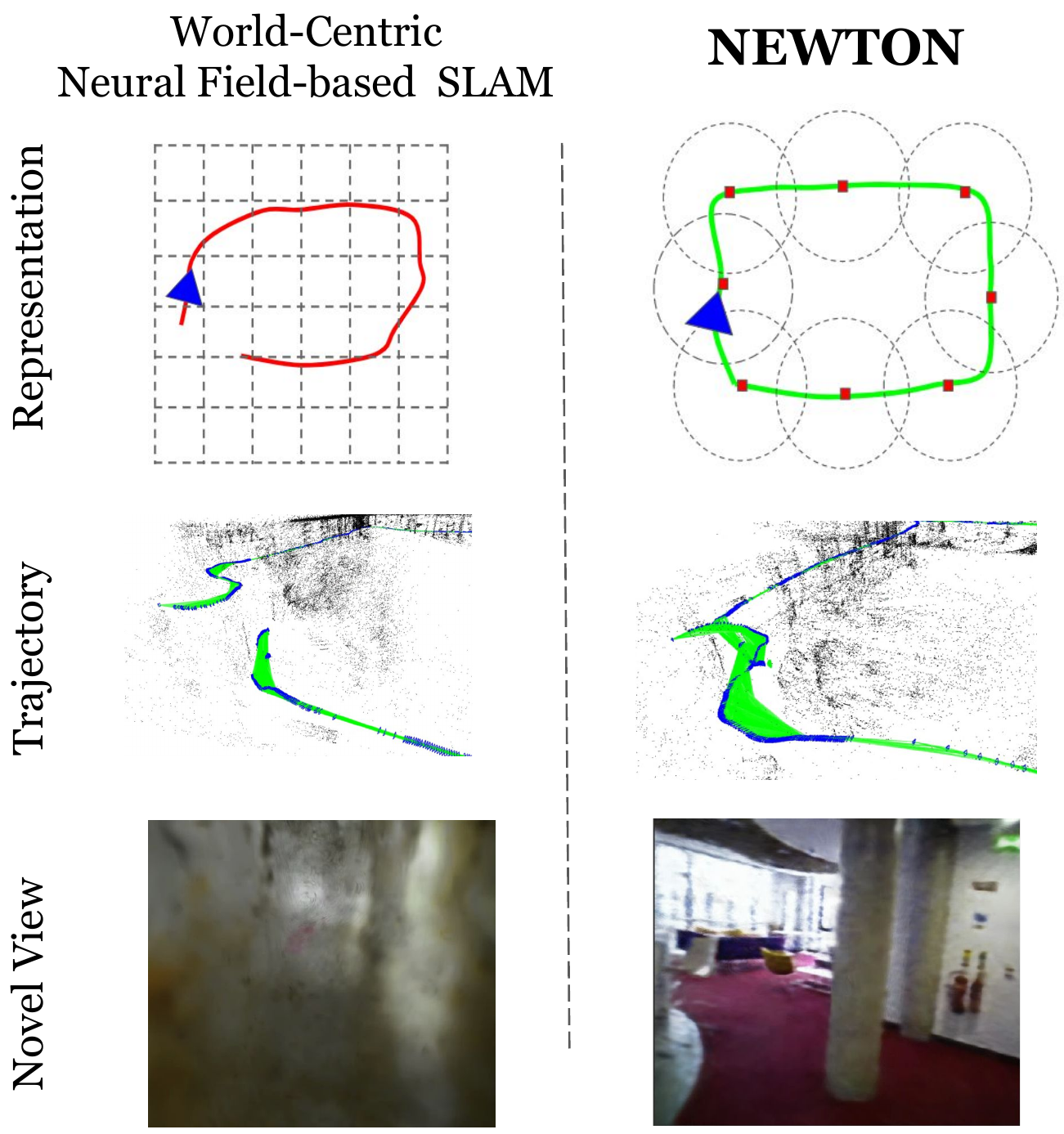}
				\caption{
    \textbf{View-Centric Mapping with NEWTON.} Prior neural field-based SLAM systems~\cite{SucaretalICCV2021,Zhu2022CVPR,Rosinol2022ARXIV} use a world-centric representation and thus cannot handle large-scale real-world scenarios where online pose updates by loop closure are required.
                In contrast, we propose a \textit{view-centric} mapping approach with multiple, local neural fields
                which are dynamically allocated.
                This improves performance and enables on-the-fly mapping for large-scale scenes.
    }
		\label{fig:tether}

\end{figure}

In recent years, neural field-based representations have enabled remarkable progress in many 3D vision applications. While initial work focused on offline 3D reconstruction~\cite{Mescheder2019CVPR,Park2019CVPR,Chen2019CVPR} and view synthesis~\cite{Sitzmann2020NIPS,Niemeyer2020CVPR,Yariv2020NIPS,Mildenhall2020ECCV}, it has now been applied to various other tasks in the area of 3D reconstruction, such as accurate and dense geometry reconstruction of large-scale scenes~\cite{Peng2020ECCV,Chibane2020CVPR,Jiang2020CVPR}, semantic fusion~\cite{Yhi2021ICCV}, and more however requiring long optimization time. Further, by leveraging an explicit volumetric representation~\cite{Mueller2022TOC,yu_and_fridovichkeil2021plenoxels}, the optimization time can be reduced from several days to operating in real-time. %

Recent work~\cite{SucaretalICCV2021, Zhu2022CVPR, Rosinol2022ARXIV} has adapted neural field-based scene representation for the task of Simultaneous Localization and Mapping (SLAM). Existing approaches rely on world-centric representations since they employ one 'global' neural field to represent the underlying scene. As a result, they have fundamental limitations for open world on-the-fly mapping, which requires loop closure and dynamic definition of scene boundaries. In particular, related work uses scene representations through a MLP~\cite{SucaretalICCV2021} or a Feature Grid~\cite{Rosinol2022ARXIV,Zhu2022CVPR} defined in a global world coordinate system and allocate features within predetermined boundaries.
These representations, however, do not allow for loop closure modules that can perform large and instant pose updates, as it either causes the tracking to fail due to the resulting inconsistent map or it requires re-training the entire field for online-mapping, since camera pose optimization tends to distort the geometry around the corrected camera positions causing inconsistencies. Especially under large-scale settings this results in misalignments and performance degradation, hence limiting the use of these methods to more controlled and small room-scale scenarios only.

In this work, we propose \textit{NEWTON}, a view-centric neural field-based representation for mapping which can handle dynamically changing input data streams in real-time. 
In contrast to prior works, we achieve this by representing the scene, as multiple neural fields defined in the local coordinate system of the respective keyframes. Further, our system is capable of dynamically allocating the individual representations based on run-time observations, which enables flexible field updates without re-training the entire scene representations.
To enable full camera pose tracking and mapping from input video stream, we combine our mapping system with the camera tracker of a loop-closable state-of-the-art SLAM system~\cite{mur2017orb}.
Our method shows superior performance and robustness on large-scale scenes.
To summarize, our contributions are as follows:
\begin{itemize}
\item A neural field-based map representation which can handle dynamic pose updates by using multiple local models.
\item A spherical parameterization for neural fields to efficiently represent unbounded scenes not limited by object-centric motion.
\item Model creation, initialization, training, and view synthesis policies for the proposed representation.

\end{itemize}

\section{Related Work}
\paragraph{Neural Fields} 
While first works on neural fields for 3D reconstruction~\cite{Mescheder2019CVPR,Park2019CVPR,Chen2019CVPR} or view synthesis~\cite{Sitzmann2020NIPS,Niemeyer2020CVPR,Yariv2020NIPS,Mildenhall2020ECCV} propose to use a single multi-layer perceptron (MLP) as scene representation, recent works propose to use spatially-distributed grid features with MLPs~\cite{Peng2020ECCV,Mueller2022TOC}, separate fore- and background models~\cite{Zhang2020ARXIVc}, space contractions~\cite{barron2022mipnerf360}, or multiple individual models~\cite{tancik2022blocknerf}. For example, BlockNeRF~\cite{tancik2022blocknerf} partitions large scenes into individual blocks and reconstructs them independently. While leading to improvements for the respective scenarios, all of these methods are world-centric, \ie they need to know a priori where to reconstruct and how to structure the scene.
While works that combine image-based rendering techniques with neural fields~\cite{Wizadwongsa2021CVPR,Wang2021CVPRc,Yu2021CVPRc} do not reconstruct in a canonical world coordinate system, they still assume a single coordinate system with aligned poses.
This, however, is not a realistic assumption for on-the-fly mapping where optimal model allocation such as uniform block partitions in the scene is impossible to define prior to the optimization.
In contrast, we propose a view-centric representation that can handle dynamically-changing input by representing scenes using multiple fields in local coordinate systems.

\paragraph{Visual SLAM}
Visual SLAM systems can be broadly classified by their underlying representation into world- and view-centric approaches.

World-centric representations assume that rich scene information such as accurate scene boundaries, \etc, are provided. 
While pioneering works such as KinectFusion~\cite{Newcombe2011ISMAR} and SuperEight~\cite{VespaRAL18} tracks against underlying voxel grid-based Truncated Signed Distance Functions (TSDF) with classical depth map fusion, later works such as NeuralRecon~\cite{sun2021neucon} and Atlas~\cite{murez2020atlas} incorporate learned priors into the volumetric TSDF mapping.
More recently, neural field-based scene representations have been incorporated due to their simplicity and state-of-the-art performance in tasks such as 3D reconstruction.
iMAP~\cite{SucaretalICCV2021} optimizes a single MLP-based scene representation from RGB-D input. NICE-SLAM~\cite{Zhu2022CVPR} improves the scene representation by combining a small MLP with spatially-distributed latent grid features, improving the mapping performance. NeRF-SLAM~\cite{Rosinol2022ARXIV} further utilizes uncertainty information given by the depth maps.
While these methods perform well for small scenes and simple camera trajectories, their performance drops significantly for real-world scenarios where dynamic pose updates and loop closure is required. 

View-centric representations in contrast construct the scene representation on a per-view basis and hence no prior global scene information such as scene boundaries or grid resolution are required for pose and map optimization. As a result, most of the state-of-the-art camera tracking systems are view-centric methods with representations such as depth maps~\cite{mur2017orb,Engel2018PAMI,Bloesch2018CVPR, teed2021droid} or surfels~\cite{Schops_2019_CVPR}.
While view-centric representations cannot directly optimize a scene globally, having multiple local maps simplifies the camera pose drift correction. This idea is powerful and some volumetric SLAM methods implement loop closure by representing a scene by multiple local chunks or submaps~\cite{dai2017bundlefusion, kahler2016real}.
As a result, in this work we propose a view-centric scene representation and show how neural fields can be used as local models, improving mapping performance over prior neural-field based approaches in particular for large-scale scenes where dynamic pose updates are required.

\section{Method}
\begin{figure*}[h]
        \centering
		\includegraphics[width=\linewidth]{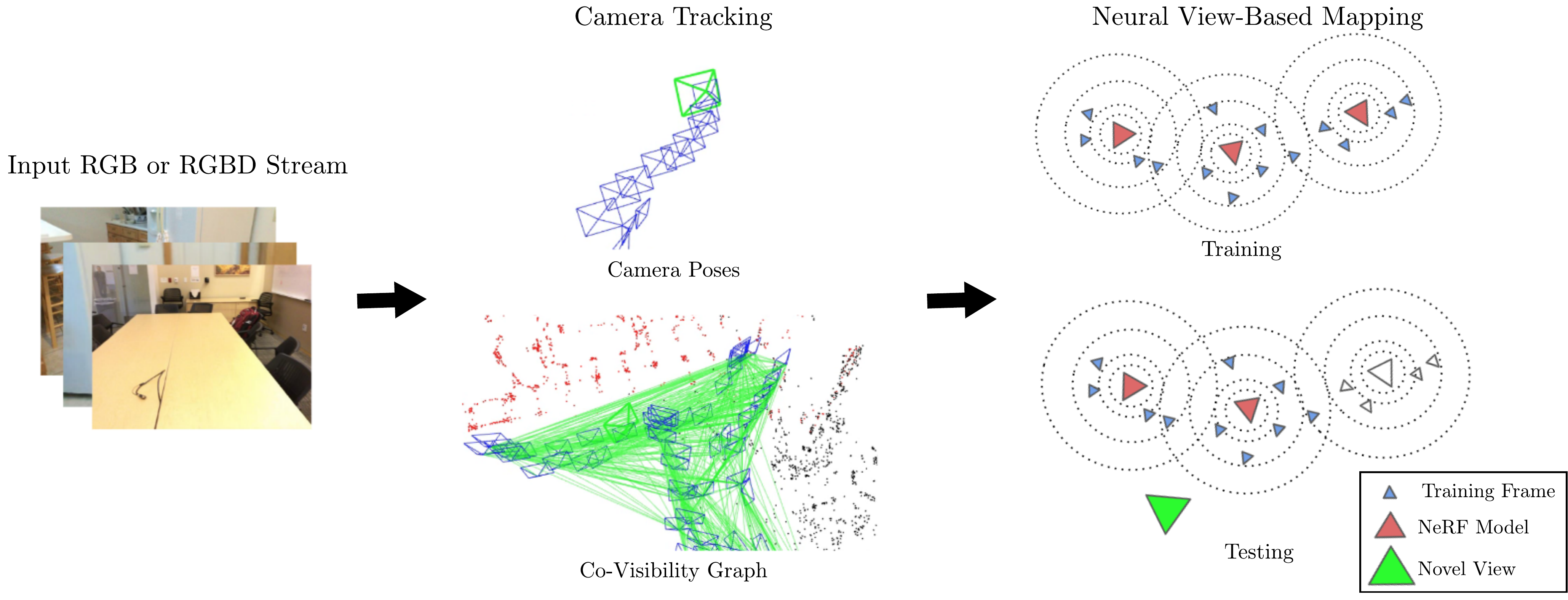}
				\caption{\textbf{Method overview.}
    The input RGB or RGB-D stream is fed to a state-of-the-art camera tracking system~\cite{mur2017orb} to obtain camera poses and a covisibility graph.
    Next, for mapping we
    train multiple neural radiance fields each defined in a local coordinate system. At test time, neighboring models are queried and images can be rendered from arbitrary viewpoints, where blending is performed in 2D image space.
    }
		\label{fig:overview}

\end{figure*}

This section introduces the proposed neural view-centric mapping approach for real-time large-scale SLAM. 
More specifically, we first obtain camera pose estimates and a co-visibility graph from an input RGB stream by means of a state-of-the-art camera tracking component~\cite{mur2017orb} (Sec.~\ref{subsec:camtrack}).
Next, for mapping we represent scenes using multiple neural radiance fields, each defined in local coordinate systems (Sec.~\ref{subsec:scenerep}).
We dynamically allocate new fields based on co-visiblity and perform feature propagation to facilitate training (Sec.~\ref{subsec:dynal}).
Fig.~\ref{fig:overview} shows an overview over our method.

\subsection{Camera Tracking}\label{subsec:camtrack}

Our proposed neural field-based mapping approach runs in combination with a separate camera tracking component. 
More specifically, the input RGB or RGB-D stream is first fed to the tracking component which returns camera poses and a co-visibility graph.
In this work, we use ORB-SLAM~\cite{mur2017orb} as it provides reliable pose estimation regardless of the scale and domain of the scene, and it performs loop-closure during runtime. 
Our mapping system then takes input keyframe poses and images as training frames and assigns them to the training view pool for the NeRF models using the frame covisibility.
Since each model is defined in a local coordinate system of the selected keyframe, it can be flexibly adjusted to the pose update and the map is much more robust than world-centric representations.

\subsection{Scene Representation}\label{subsec:scenerep}

\subsubsection{Spherical Multi-Resolution Hash Encoding}
\begin{figure}[h]
        \centering
		\includegraphics[width=\linewidth]{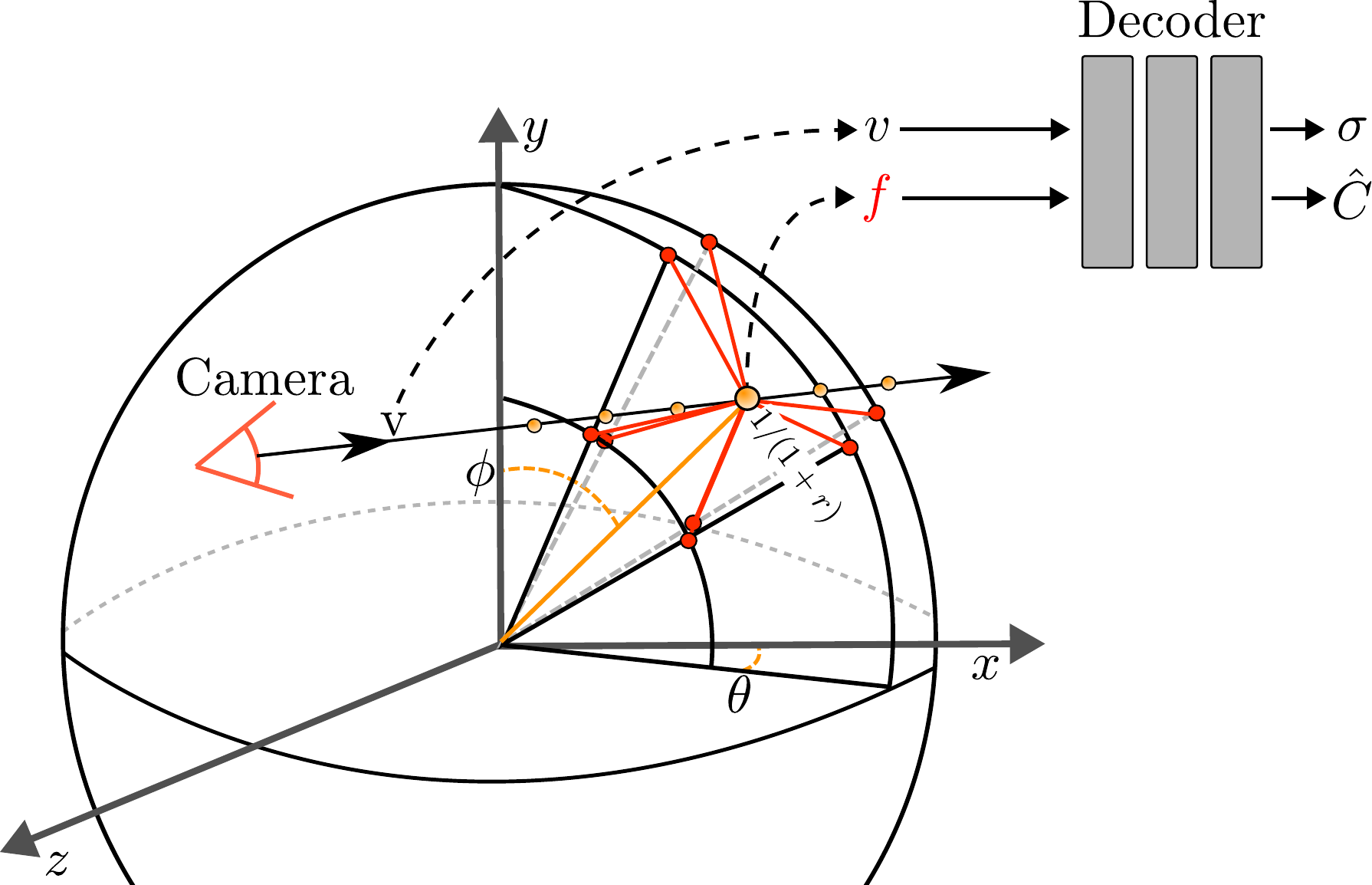}
				\caption{\textbf{Spherical Multi-Resolution Grid.} This representation allocates features evenly along inverse distance and naturally supports unbounded scene without a limitation of the field of view.}
		\label{fig:shpericalgrid}
\end{figure}

We represent scenes with multiple NeRF models defined in the local coordinate system of selected keyframes. 
We represent each NeRF model as a multi-resolution feature grid~\cite{Mueller2022TOC} in a spherical coordinate system (see~Fig.~\ref{fig:shpericalgrid}).
We convert the input location represented in Cartesian coordinates ${(x, y, z)} \in \mathbb{R}^3$ to spherical coordinates ${(\phi, \theta, \frac{1}{1+r})} \in [0, 1]^3$ 
as follows:
\begin{align}
\begin{split}
    r &= \sqrt{x^{2} + y^{2} + z^{2}} \\
    \theta &= \frac{2}{\pi} \arctan\left(\frac{z}{x}\right)  \\
    \phi &= \frac{1}{\pi} \arcsin{\frac{y}{r}}.
\end{split}
\end{align}
This enables at the same time an efficient feature allocation strategy related to the pixel's measurement accuracy and a representation that is not limited by the frustum boundary.

\begin{figure}[h]
        \centering
            \begin{subfigure}[b]{.495\linewidth}
            \includegraphics[width=\linewidth]{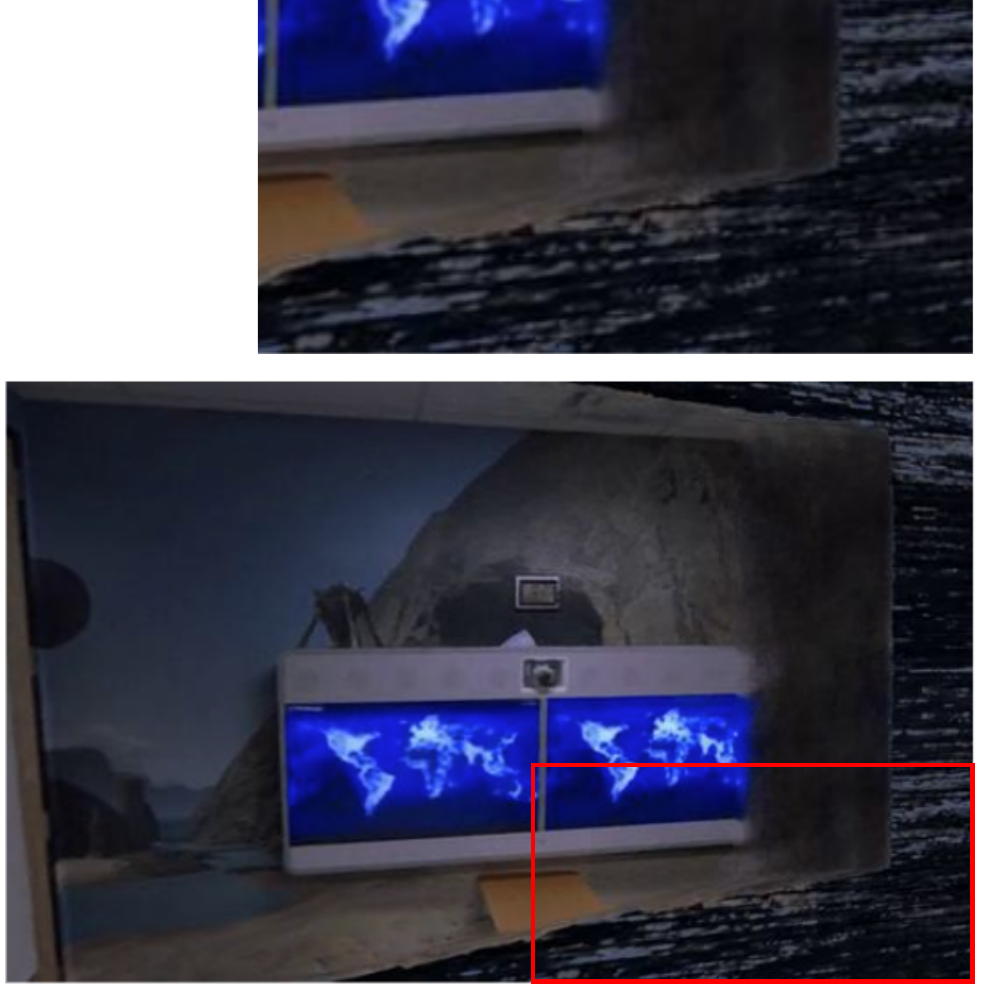}
            \caption{\small NDC Parameterization}
        \end{subfigure}
            \begin{subfigure}[b]{.495\linewidth}
            \includegraphics[width=\linewidth]{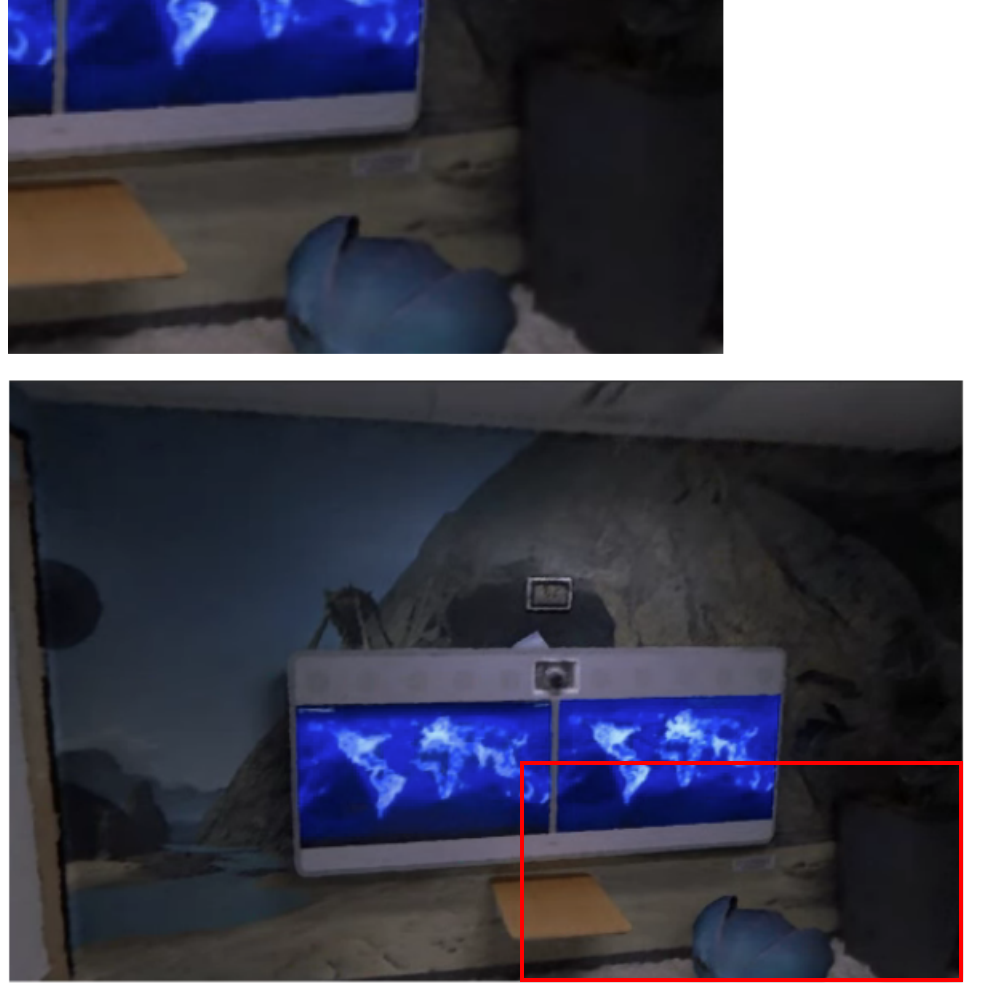}
            \caption{\small Our Parameterization}
        \end{subfigure}
				\caption{\textbf{NDC vs Spherical Parameterization.}
    While both parameterizations allocate features along inverse depth, NDC lead suffers from artifacts around the boundaries. In contrast, our proposed spherical parameterization leads to high-quality view synthesis results for the entire image.
    }
		\label{fig:ndc_vs_ours}
\end{figure}

\paragraph{Feature Allocation}
An increasing distance to the camera center leads to a larger measurement uncertainty as disparity is inversely-proportional to its depth~\cite{civera2008inverse}.
To allocate more capacity in areas of high certainty, it is therefore efficient to allocate the features evenly in inverse depth space.
This representation can be interpreted as an omnidirectional extension of a Normalized Device Coordinates (NDC) used for novel view synthesis in forward facing scenes in ~\cite{Mildenhall2020ECCV}. 
While the NDC parameterization contracts the scene along inverse depth, the region is strictly bounded by the model view frustum, which discards the training view information outside of the frustum and also causes artifacts around the frustum boundary as shown in Fig.~\ref{fig:ndc_vs_ours}. In contrast, our spherical representation can make full use of the training view information without being limited to the field of view.

\paragraph{Unbounded Scenes}
Our representation can naturally capture unbounded scenes.
While other parameterizations such as the recently proposed space contraction from~\cite{barron2022mipnerf360}
can also represent $360^\circ$ unbounded scenes, it explicitly separates foreground and background. This is effective only when the camera trajectory is object-centric and inward-looking, and the region of interest needs to be known a priori in world coordinates. However, in case of general odometry-type video sequences, the camera trajectory is not limited to being object-centric and it is not possible to define the region of interest beforehand. 

\subsubsection{Loss Functions}

We use classic volume rendering techniques~\cite{Mildenhall2020ECCV,Mueller2022TOC} to render pixel color $\hat{C}$. We train each NeRF model by minimizing the MSE loss against ground truth color ${C}$:
\begin{eqnarray}
L_\text{rgb} = \sum_{r \in \mathcal{R}} \vert\vert\hat{C_r} - {C_r}\vert\vert_2^2.
\end{eqnarray}
where $r \in \mathcal{R}$ are rays subsampled from all pixels/rays.
We further incorporate a distortion regularization~\cite{barron2022mipnerf360}. 
Let $w_s(u) = \sum{i}w_i\mathbb{1}_{s_i, s_{i+1}}(u)$ be interpolation into the step function, a distortion regularization is defined as:
\begin{align}
\begin{split}
L_\text{dist} = &\sum_{i,j} {w{_i}w{_j}} {\left|{\frac{s_i + s_{i+1}}{2} - \frac{s_j + s_{j+1}} {2}}\right|} \\ &+ \frac{1}{3}\sum_{i} {w_i^{2}(s_{i+1} - s_{i})}.
\end{split}
\end{align}
We also follow~\cite{barron2022mipnerf360} and add a small proposal network for each NeRF model. Here we define ray interval and predicted weight vector from the main NeRF network as $\bf{t}, \bf{w}$ and those from the proposal network as $\hat{\bf{t}}, \hat{\bf{w}}$. The bound function with interval $T$ is defined as:
\begin{eqnarray}
\text{bound}(\hat{\bf{t}}, \hat{\bf{w}}, T) = \sum_{j:T\cap\hat{T_j}\neq\varnothing}\hat{w}_j.
\end{eqnarray}

We train the proposal network with the following proposal loss,
\begin{eqnarray}
L_\text{prop}(t, w,\hat{t},\hat{w}) = \sum_{i}\max(0, w_i-\text{bound}(\hat{t},\hat{w},T_i))^2
\end{eqnarray}

We can optionally incorporate depth measurement if available. Similar to a pixel's color value, we use volume rendering to obtain a pixel's rendered $\hat{D}_r$ and minimize
\begin{eqnarray}
L_\text{depth} = \sum_{r \in \mathcal{R}} \vert\hat{D}_r - {D}_r\vert_1
\end{eqnarray}
where ${D}$ is the GT depth for ray $r$.
The full loss we optimize is given by:
\begin{align}\begin{split}
 L_\text{total} = L_\text{rgb} + \lambda_\text{dist}L_\text{dist} + \lambda_\text{prop}L_\text{prop} +\lambda_\text{depth}L_\text{depth}.
\end{split}\end{align}

\subsection{Dynamic Neural Field Allocation and Training}\label{subsec:dynal}

\subsubsection{Training Policy}
When a new keyframe is created in our camera tracking module, we check for covisibility between the keyframe and the existing training frames of NEWTON by using the covisibility graph. Each training frame is assigned to a single model ID which represents the primary model it belongs to. We check the model IDs of the covisible training frames, assign the new keyframe to the newest covisible model, and further assign this newest model ID to the frame. We also add the new keyframe to the training batch of the other older covisible models to keep appearance consistency of spatially-overlapping regions for different models. For every iteration, we train up to three models simultaneously; two of the newest covisible models from the latest training frame and one randomly chosen model.

\subsubsection{Model Creation}\label{subsubsec:modelcreation}
\paragraph{Creation Policy}
If the input keyframe is further away than a distance threshold $d_{th}$ from any of the covisible models, we create a new model. To better initialize the new model we apply our proposed feature propagation.

\paragraph{Feature Propagation}
\begin{figure}
        \centering
		\includegraphics[width=\linewidth]
  {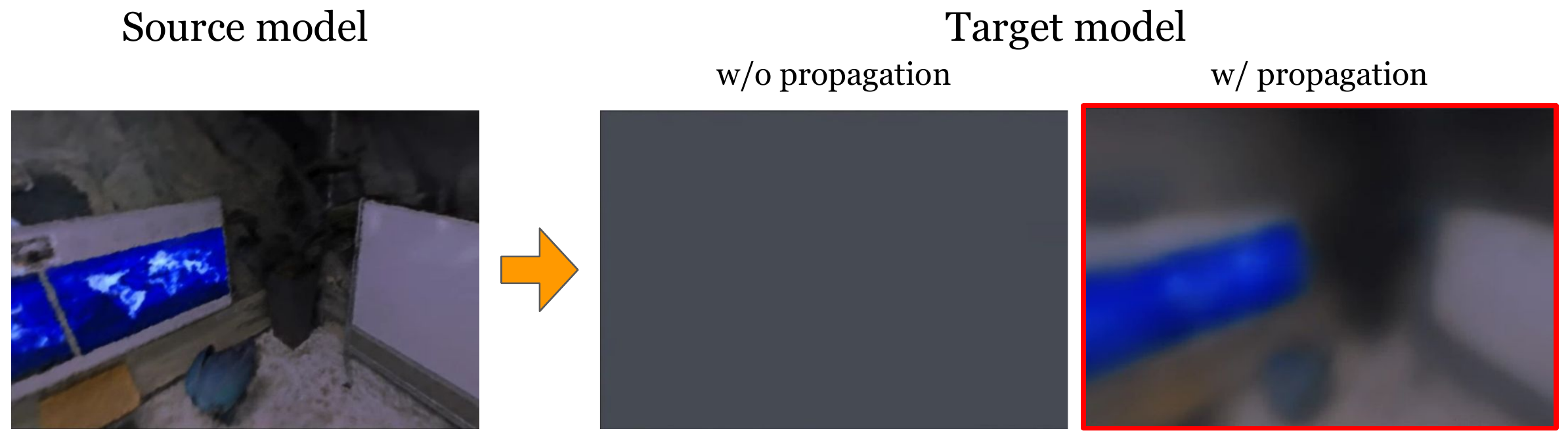}
				\caption{\textbf{Feature Propagation.} We initialize new models by propagating parameters of previous models
                which improves both photometric and geometric accuracy.
    }
		\label{fig:feature_propagation}

\end{figure}

When creating a new model, it is better to initialize the model parameters using the information from previously trained models instead of training from scratch. Here we describe how to propagate the trained feature information from the previous model to the new one. For the coarse level of multi-resolution hash grids which does not have any hash collision, we query the copy source model and trilinearly interpolate the feature on each corner of the new feature grid. We do not propagate the fine level features from the hash table because the hash function is uniquely defined in one local coordinate and the features are learned to handle collisions in the specific coordinate system. This cannot be easily transferred to another coordinate system. We use the weights of the MLPs as a coarse initialization of the newly created field. We investigate the effect of the feature propagation in an ablation study.

\subsubsection{Novel View Rendering}
For novel view synthesis, we first perform model selection based on the camera position and query multiple NeRF models. We then perform image blending in 2D image space.

\paragraph{Model Selection}
For a novel view, we first check the nearest training view position, then take the top three closest models for this training view. We render an image for each model and blend them. %
For each model, we also store the coarse voxel grid (of resolution ${16}^3$) that contains information whether a block in the grid is used during training. For rendering at test time, we check the visibility for queried sample points and perform empty space skipping.

\paragraph{Blending}
We blend rendered images by weighting them based on the distance between the novel view and the nearest training view of each queries model. Let the camera origin of the test view be $o$ and $t_i$ the position of the nearest training view for the $i$-th model. Similar to~\cite{tancik2022blocknerf}, we use the inverse distance weighting function:
\begin{align}
w_i \propto \vert\vert o - t_i \vert\vert_2^{-p}.
\end{align}

\section{Evaluation}
\subsection{Experimental Setup}

We first evaluate our method for \textbf{on-the-fly mapping} and investigate its performance when dynamic pose updates are required. 
Next, we run a \textbf{baseline comparison} against the following state-of-the-art neural field-based SLAM systems: NeRF-SLAM~\cite{Rosinol2022ARXIV}, NICE-SLAM~\cite{Zhu2022CVPR}, and iMAP~\cite{SucaretalICCV2021}. 
Finally, we perform an \textbf{ablation study} of various components of our method.

\paragraph{Implementation Details}

We train our method on a single NVIDIA RTX3090Ti GPU which can process our system with a 20 fps video stream input.
We build our method on an unbounded instant-NGP implementation in Nerfstudio~\cite{nerfstudio} and use the same hash size and ray parameters. We set  $\lambda_{dist}$=0.002 , $\lambda_{prop}$=1.0, $\lambda_{depth}$=0.5, $p$=4, $d_{th}$=0.3.

\paragraph{Datasets}
We use in total seven real-world scenes where ORB-SLAM~\cite{mur2017orb} detects loop closure, ranging from small room-scale to large building-scale sequences. We take one sequence from Kintinuous~\cite{Whelan14ijrr} recorded by a hand-held Kinect v1 camera in a large building which has a large loop closure at the end of the sequence.
We further follow previous works~\cite{Zhu2022CVPR} and take six sequences from the ScanNet dataset~\cite{dai2017scannet} which contains small to middle-sized room sequences recorded by Kinect v1. While we run the input video stream at 20Hz for our method, the baseline methods run offline and process frames sequentially. 

\paragraph{Metrics}
We report reconstruction metrics of unseen views \wrt both, the predicted appreance a well as the geometry. For the first, we report PSNR, SSIM, and LPIPS, while for measuring the predicted geometry, we report the L1 metric on predicted and GT depth maps in meter scale.

\subsection{On-the-fly Mapping}
\paragraph{Experiment Design}
We evaluate our on-the-fly mapping performance by comparing NEWTON to a world-centric variant of our approach with a single NeRF model which we call World-Centric Single Model.
We use the ORB-SLAM's trajectory from RGBD as input for both representations.

We use 10\% of ORB-SLAM's localized keyframes as test frames and exclude them from training.
To measure temporal run-time performance, we report reconstruction metrics of all test views for a certain interval (every 50 keyframe creation by ORB-SLAM).
For each dataset, we run NeRFStudio~\cite{nerfstudio}'s preprocessing pipeline and rescale the scene to fit the default model and ray parameters.

\paragraph{Result}

\setlength{\tabcolsep}{1.0mm}
\begin{table*}
\centering
     \resizebox{\linewidth}{!}{
    \begin{tabular}{p{3em}|c|c|cc|cc|cc|cc|cc|cc|cc|cc|}
    \hline
    \multicolumn{3}{|c|}{}& \multicolumn{2}{c|}{Kintinuous} &  \multicolumn{12}{c|}{ScanNet} & \multicolumn{2}{c|}{} \\
    \hline
      \multicolumn{3}{|c|}{}& \multicolumn{2}{c|}{loop} & \multicolumn{2}{c|}{0000} & \multicolumn{2}{c|}{0059} & \multicolumn{2}{c|}{0169} & \multicolumn{2}{c|}{0181} & \multicolumn{2}{c|}{0207} & \multicolumn{2}{c|}{0106} & 
      \multicolumn{2}{c|}{Avg.} \\
    \multicolumn{3}{|c|}{}& Mean & Std & Mean & Std & Mean & Std & Mean & Std & Mean & Std & Mean & Std & Mean & Std & Mean & Std    \\
    \hline\hline
    \multicolumn{2}{|c|}{\begin{tabular}{c}World-Centric \\ Single Model \end{tabular} }
    &\begin{tabular}{c} PSNR$\uparrow$\\ SSIM$\uparrow$ \\ LPIPS$\downarrow$ \\ L1Depth$\downarrow$\end{tabular} 
    &\begin{tabular}{c}20.01\\0.654 \\0.204 \\ 0.364\end{tabular} &\begin{tabular}{c}2.021\\0.079 \\0.038 \\0.211\end{tabular} 
    
    &\begin{tabular}{c}23.36\\0.765 \\\bf0.098 \\\bf0.187 \end{tabular} &
    \begin{tabular}{c}1.243\\0.059 \\ 0.014\\ 0.010\end{tabular} 
    
    &\begin{tabular}{c}\bf19.40\\\bf0.646 \\  \bf0.159 \\ 0.267 \end{tabular}
    &\begin{tabular}{c}1.673 \\0.104 \\ 0.016 \\ 0.035\end{tabular}
    
    &\begin{tabular}{c}24.13\\0.714 \\\bf0.167 \\ 0.217 \end{tabular}
    &\begin{tabular}{c}2.551\\0.074 \\0.024 \\ 0.035\end{tabular} 
    
    &\begin{tabular}{c}21.49\\0.707 \\ 0.139\\ 0.199\end{tabular} 
    &\begin{tabular}{c}2.002\\0.062 \\ 0.029\\ 0.026\end{tabular} 
    
    &\begin{tabular}{c}22.21\\0.676 \\ \bf0.176 \\ 0.200 \end{tabular} 
    &\begin{tabular}{c}2.278\\0.087 \\ \bf0.024\\ 0.025\end{tabular}

    &\begin{tabular}{c}\bf20.35\\\bf0.686 \\ \bf0.164 \\ \bf0.247\end{tabular}
    &\begin{tabular}{c} \bf0.659\\0.033 \\ \bf0.018 \\ \bf0.016\end{tabular}
    
    &\begin{tabular}{c}21.57 \\0.692 \\ 0.176 \\  0.240\end{tabular}
    &\begin{tabular}{c}1.770\\0.071 \\ 0.023 \\ 0.051\end{tabular}
    \\
    \hline

    \multicolumn{2}{|c|}{Ours}
    &\begin{tabular}{c} PSNR$\uparrow$\\ SSIM$\uparrow$ \\ LPIPS$\downarrow$ \\ L1Depth$\downarrow$\end{tabular} 
 &\begin{tabular}{c} \bf20.45 \\ \bf 0.684 \\\bf 0.174 \\ \bf0.311 \end{tabular} 
 &\begin{tabular}{c} \bf0.96 \\ \bf0.037 \\ \bf0.014 \\ \bf0.066 \end{tabular} 

 &\begin{tabular}{c}\bf23.58\\\bf0.773 \\0.100\\0.188\end{tabular} &\begin{tabular}{c}\bf0.772 \\\bf0.036 \\ \bf0.012 \\ \bf0.007\end{tabular} 
 
 &\begin{tabular}{c}19.36 \\0.631 \\ 0.213 \\ \bf0.264\end{tabular}
&\begin{tabular}{c}\bf0.806 \\\bf0.036 \\ \bf0.008 \\ \bf0.010\end{tabular}

    &\begin{tabular}{c}\bf24.95 \\\bf0.720 \\ 0.186 \\ \bf0.215\end{tabular}
        &\begin{tabular}{c}\bf0.643 \\\bf0.032 \\ \bf0.019 \\ \bf0.021 \end{tabular} 
    &\begin{tabular}{c}\bf22.09\\\bf0.717 \\ \bf0.138\\\bf0.193\end{tabular} 
    &\begin{tabular}{c}\bf0.627 \\\bf0.019 \\ \bf0.015 \\\bf0.016 \end{tabular} 
    
    &\begin{tabular}{c}\bf23.15 \\\bf0.702 \\ 0.183\\ \bf0.184\end{tabular} 
    &\begin{tabular}{c}\bf0.967 \\\bf0.036 \\ 0.028 \\ \bf0.005\end{tabular}
    
    &\begin{tabular}{c}20.16 \\0.683 \\ 0.178 \\ 0.274 \end{tabular}
    &\begin{tabular}{c}\bf0.659 \\0.029 \\  \bf0.018\\ 0.059\end{tabular}

    &\begin{tabular}{c}\bf21.96 \\\bf0.701 \\ \bf0.167\\ \bf0.233\end{tabular}
    &\begin{tabular}{c}\bf0.777 \\\bf0.032 \\ \bf0.016\\ \bf0.026\end{tabular}
    
    \\
    \hline    

    \end{tabular}
    }

    \caption{\textbf{Reconstruction Results for On-the-Fly Mapping.} For each scene, we report the average over the full time series.}
    \label{table:onthefly}

\end{table*}
\begin{figure}[h]
        \centering
		\includegraphics[width=\linewidth]{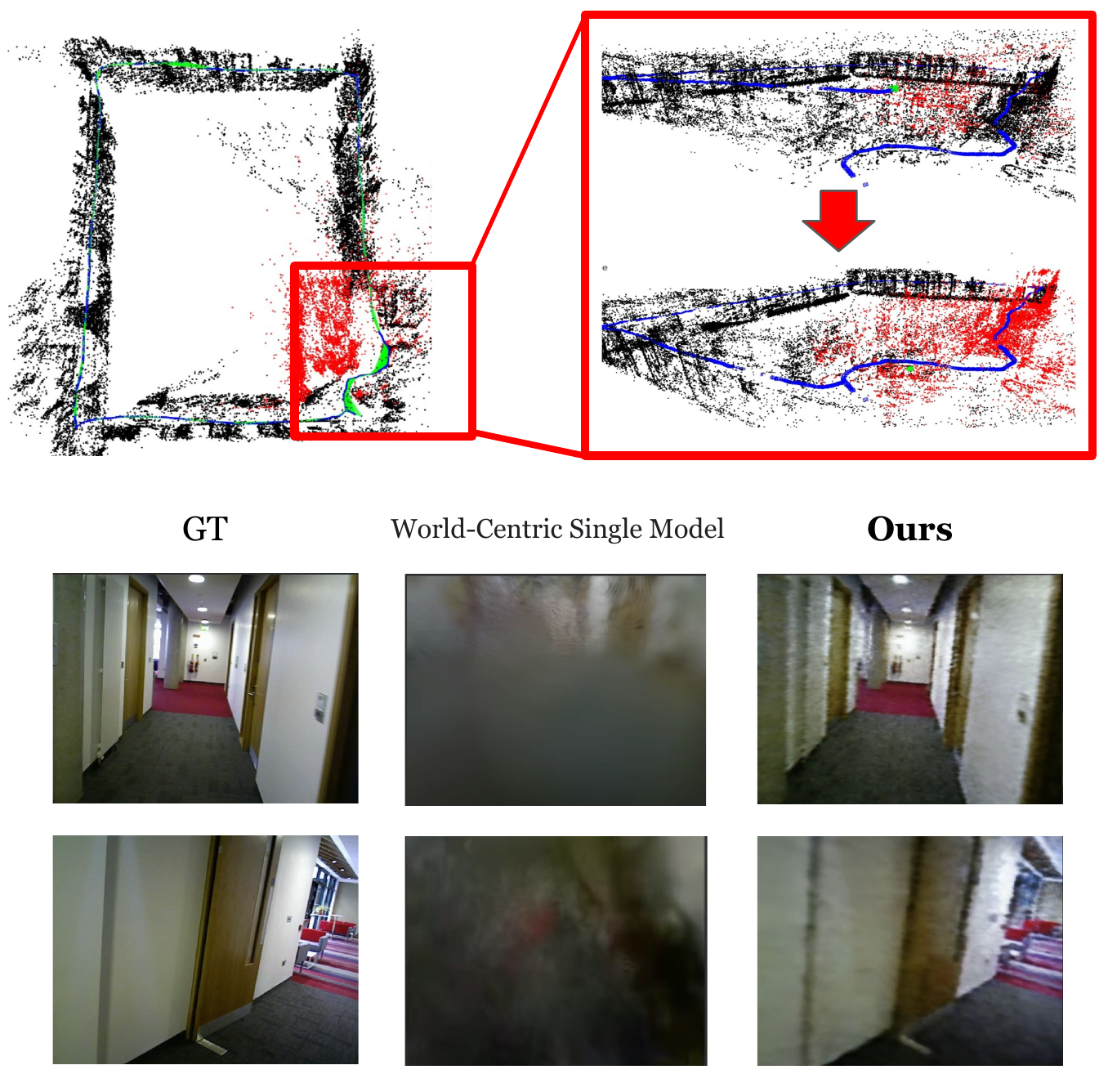}
				\caption{\textbf{Loop Closure.}
    While for a world-centric single model loop closure leads to degraded view synthesis, our view-centric approach with multiple local models allows for high-quality view synthesis even when performing loop closure (here shown for the Kintinuous dataset).
    }
		\label{fig:onthefly_kintinuous}

\end{figure}
\begin{figure}
    \centering
    \begin{subfigure}[b]{1.\linewidth}
        \includegraphics[width=0.49\linewidth]{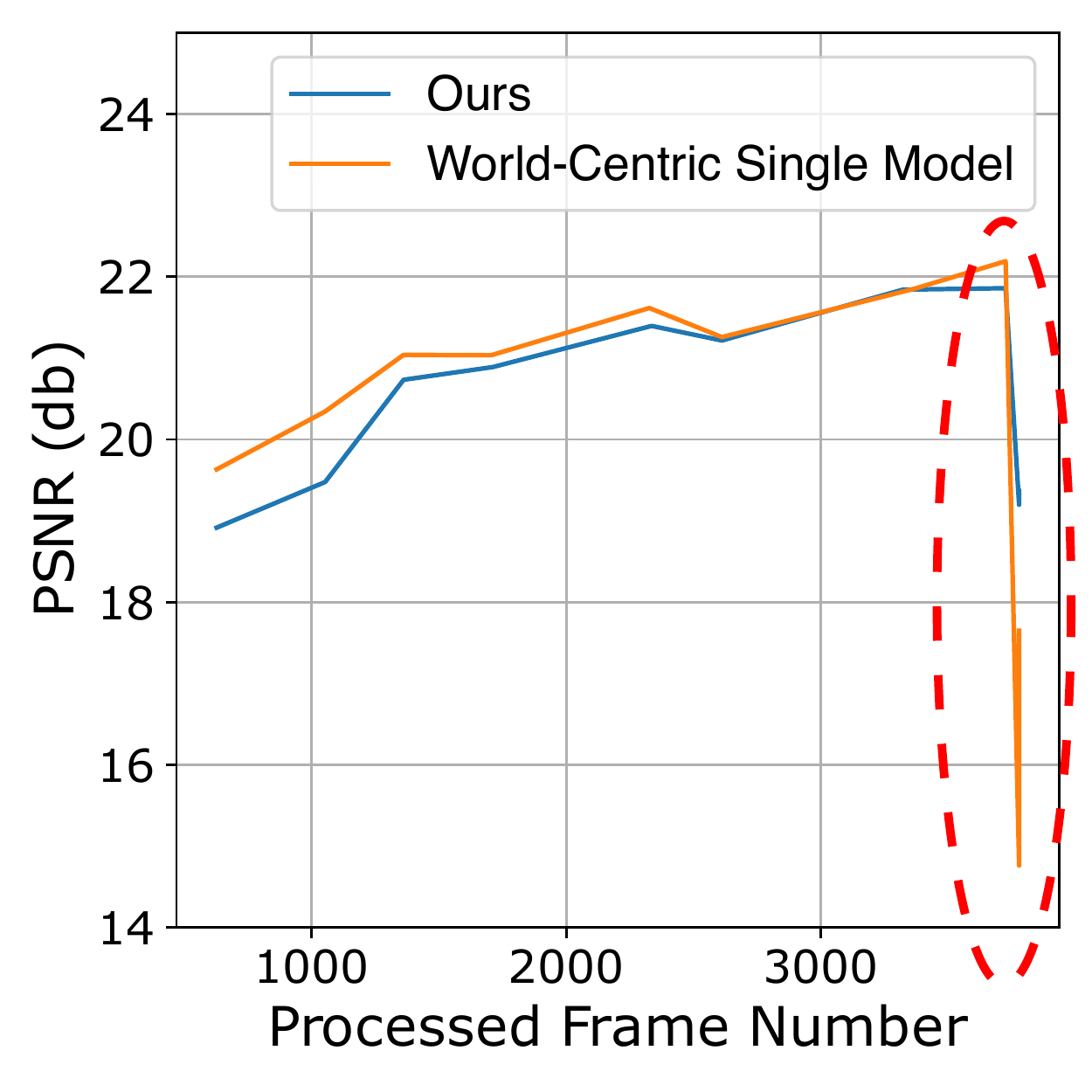}
        \includegraphics[width=0.49\linewidth]{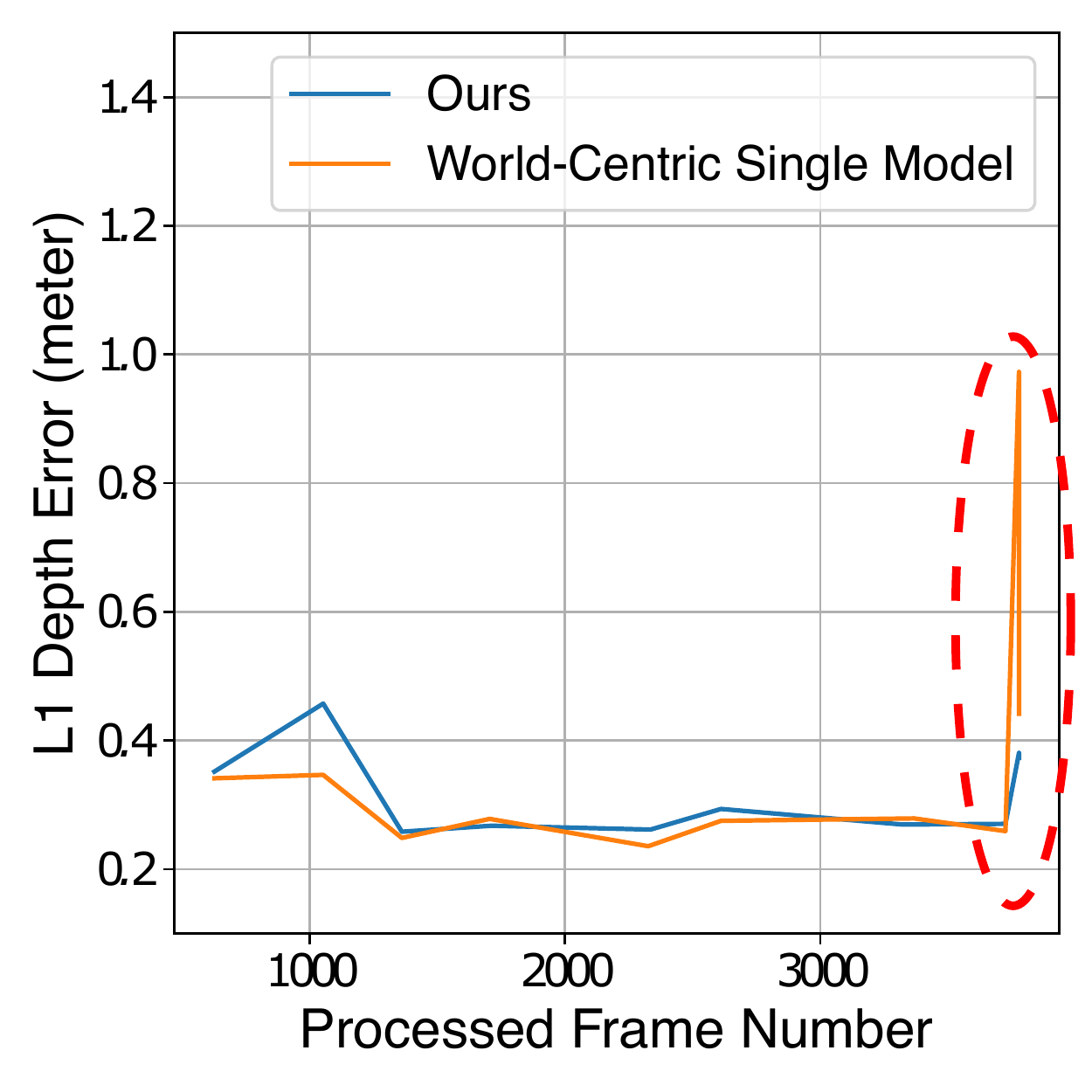}
        \caption{Kintinuous Dataset.}
    \end{subfigure}
    \begin{subfigure}[b]{1.\linewidth}
        \includegraphics[width=0.49\linewidth]{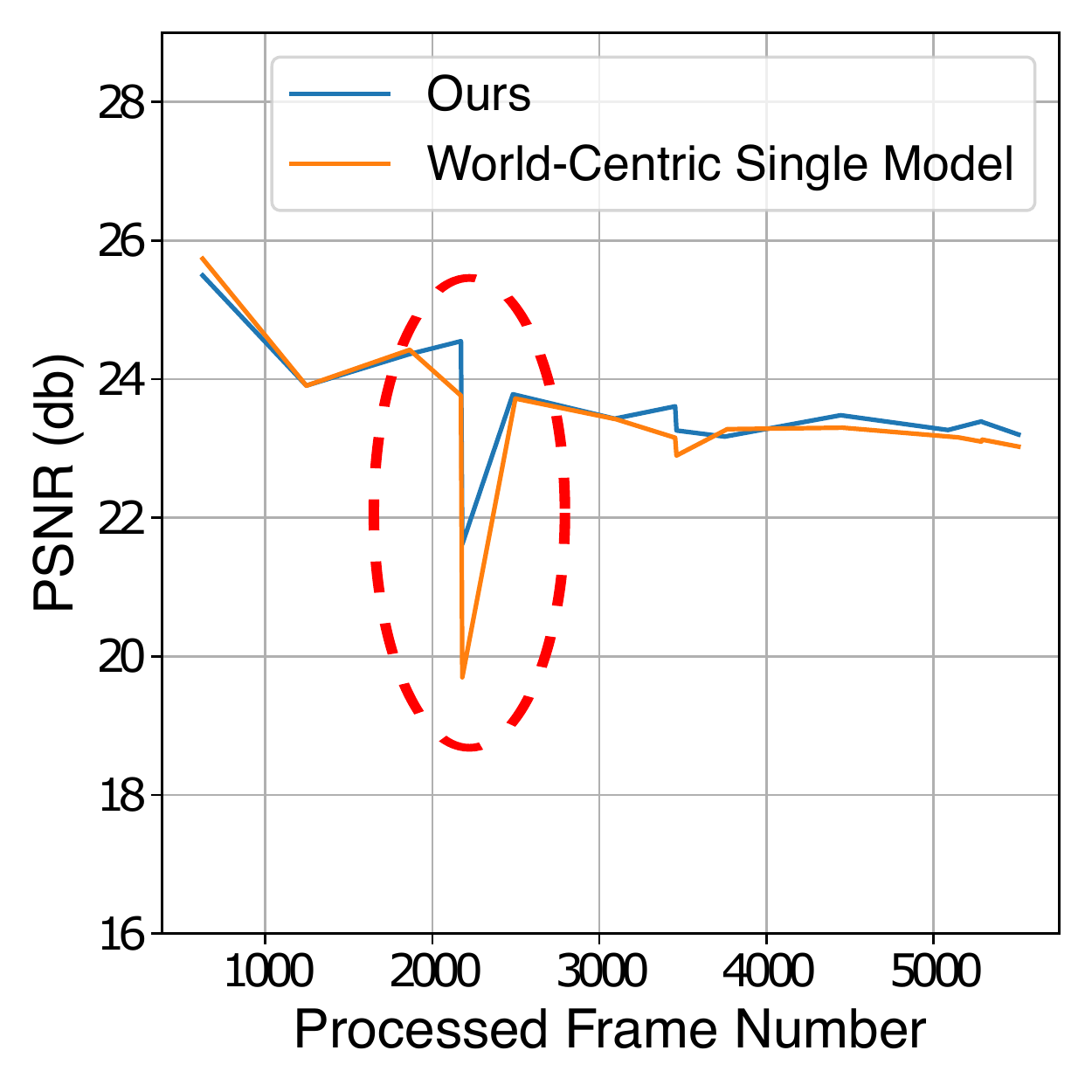}
        \includegraphics[width=0.49\linewidth]{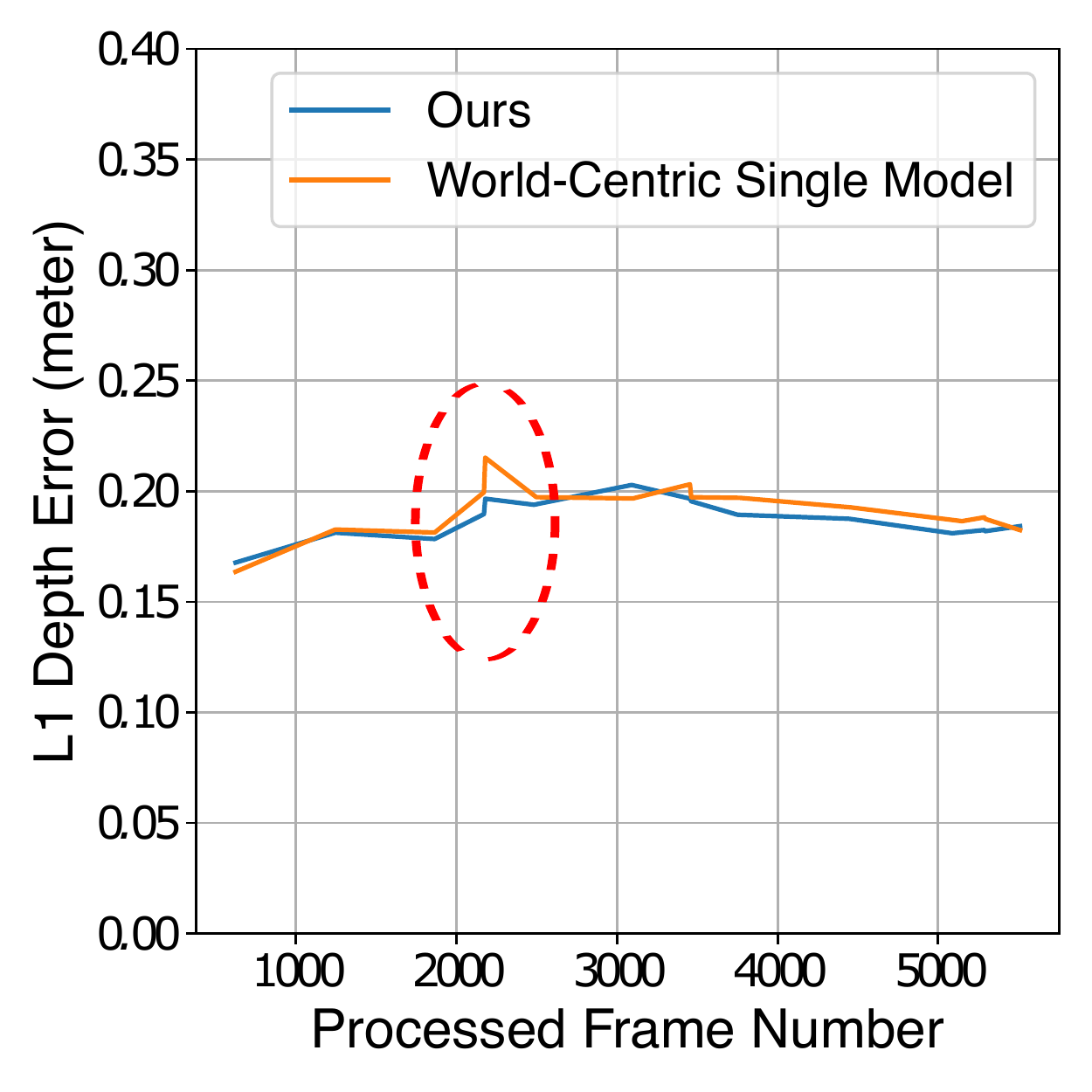}
        \caption{ScanNet Dataset (scene0000).}
    \end{subfigure}
    \caption{\textbf{Timeseries For On-the-Fly Mapping.}
    Our method is more robust to loop closure while the baseline World-Centric Single Model shows a significant performance drop (see red circles for where loop closure appears).}
    \label{fig:timeseries}
\end{figure}

The quantitative results in Table~\ref{table:onthefly} and qualitative results in Fig.~\ref{fig:onthefly_kintinuous} show that our method is minimally affected by loop closure while for the World-Centric Single Model, the performance drops significantly. Further, the time series plot in Fig.~\ref{fig:timeseries} shows stronger robustness of our approach to the dynamic pose update and lower deviation values compared to the baseline.

\begin{figure*}[h]
        \centering
		\includegraphics[width=\linewidth]{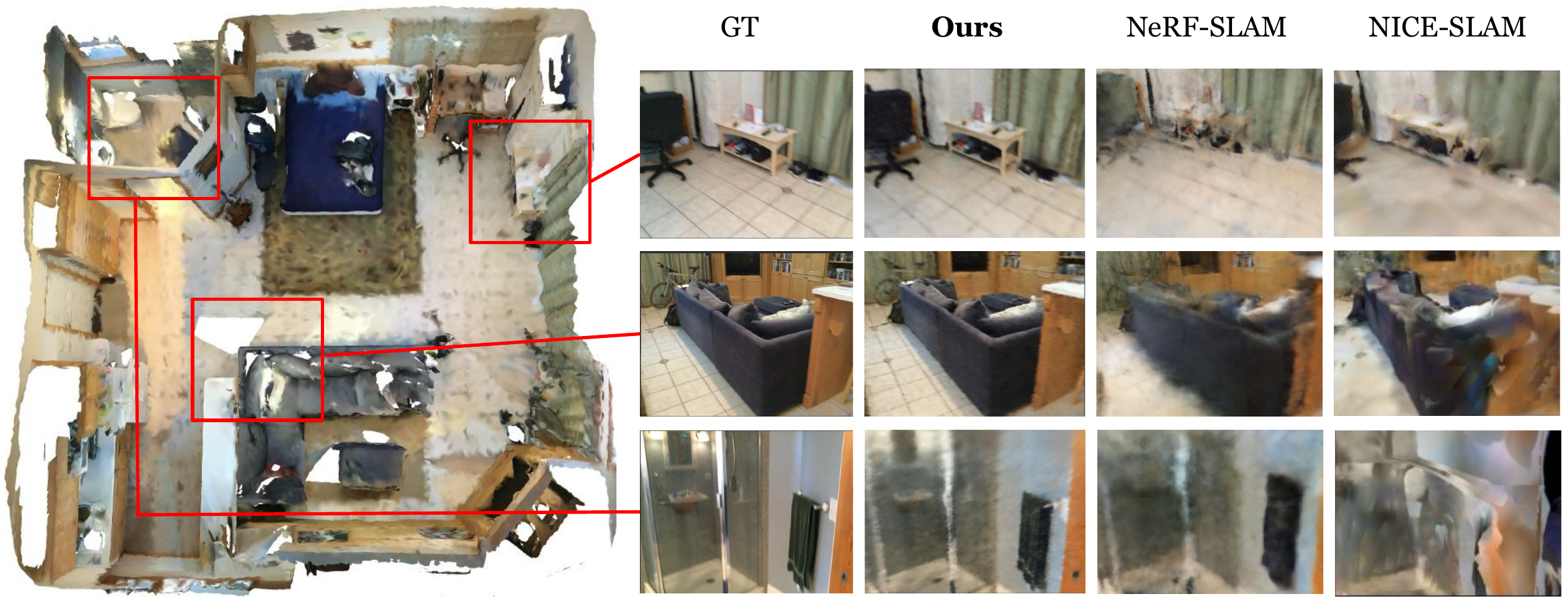}
				\caption{\textbf{View Synthesis on ScanNet (scene0000).} Left: GT mesh. Right: Novel view synthesis results.
                Existing world-centric SLAM methods such as NeRF-SLAM~\cite{Rosinol2022ARXIV} or NICE-SLAM~\cite{Zhu2022CVPR} cannot handle loop closure and show misalignment in rendered views.
                In contrast, our representation flexibly adjusts to the pose update and leads to high-quality view synthesis.
                }
		\label{fig:qualitative_scannet}

\end{figure*}

\subsection{Baseline Comparison}
\paragraph{Experiment Design}
We compare our method to the state-of-the-art neural field based SLAM methods NeRF-SLAM~\cite{Rosinol2022ARXIV}, NICE-SLAM~\cite{Zhu2022CVPR}, iMAP*~\cite{SucaretalICCV2021}~\cite{Zhu2022CVPR}. 
As these are world-centric approaches, the drift correction by loop closure is not performed and shows a clear trajectory drift. NeRF-SLAM uses DROID-SLAM~\cite{teed2021droid} with its own optimization backend and it does not perform drift correction by loop closure on these test sequences. We provide sensor depth measurements with NeRF-SLAM's frontend tracking system for fair comparison.  To evaluate the effect of the drift, we first align each trajectory to ground truth, and then render the images on the associated ground truth frame position. For large-scale scenes, it is difficult to get a perfect ground truth trajectory by using a motion capture system. In our settings, Kintinuous dataset does not provide ground truth trajectory data and the ScanNet dataset only provides trajectories estimated from BundleFusion~\cite{dai2017bundlefusion} which are less accurate than ORB-SLAM's trajectories because the sensor (Kinect v1) is not perfectly synchronized and has the rolling shutter effect~\cite{Schops_2019_CVPR}. Therefore, we adopt the frame trajectory from RGBD ORB-SLAM as ground truth for evaluation. We compare the final reconstruction result of these methods.

\paragraph{Result}
Table~\ref{table:comparison} shows the quantitative evaluation of our baseline comparison. 
We observe that our method achieves better performance for the majority of sequences. This indicates the importance of our view-centric mapping approach allowing for online drift correction and online camera pose updates when optimizing the neural field-based mapping representation. We further show qualitative results in Fig.~\ref{fig:qualitative_scannet} highlighting the effect of the trajectory drift. Notably, although our method and NeRF-SLAM both use volumetric NeRF as an underlying neural field representation, our view synthesis quality is significantly better due to the training camera pose accuracy. These results demonstrates the limitations of existing world-centric neural field-based mapping approaches for large scenes and long trajectories as well as the necessity of scene representations to be able to handle online state updates.

\begin{table*}[h]
\centering
    \begin{tabular}{p{3em}|c|c|c|c|c|c|c|c|c|c|c|}
    \hline
    \multicolumn{3}{|c|}{}& \multicolumn{1}{c|}{Kintinuous} &  \multicolumn{6}{c|}{ScanNet}& \\
    \hline
    \multicolumn{3}{|c|}{}& loop & 0000 & 0059 & 0169 & 0181 & 0207 & 0106 & Avg. \\
    \hline\hline
    \multicolumn{2}{|c|}{Ours }
    &\begin{tabular}{c} PSNR$\uparrow$\\ SSIM$\uparrow$ \\ LPIPS$\downarrow$ \\ L1Depth$\downarrow$\end{tabular} & 
 \begin{tabular}{c}\bf19.85 \\\bf0.634 \\ \bf0.193\\ \bf0.376 \end{tabular} &
 \begin{tabular}{c}\bf23.46\\\bf0.769 \\\bf0.108 \\\bf0.184 \end{tabular} &
 \begin{tabular}{c}\bf19.62\\\bf0.644 \\ \bf0.211\\ \bf0.264 \end{tabular} &
 \begin{tabular}{c}\bf24.34\\\bf0.698 \\ \bf0.198\\ \bf0.223\end{tabular} &
 
\begin{tabular}{c}\bf21.04 \\\bf0.686 \\ \bf0.162 \\ \bf0.214 \end{tabular}

    &\begin{tabular}{c}\bf22.94 \\\bf0.693 \\ \bf0.178 \\ 0.183 \end{tabular}
    &\begin{tabular}{c}\bf20.35 \\\bf0.697 \\ \bf0.173 \\ 0.294\end{tabular}
        &\begin{tabular}{c}\bf21.66 \\\bf0.689 \\ \bf0.175 \\ \bf0.248\end{tabular}

    \\
    \hline

    \multicolumn{2}{|c|}{NeRF-SLAM~\cite{Rosinol2022ARXIV}}
    &\begin{tabular}{c} PSNR$\uparrow$\\ SSIM$\uparrow$ \\ LPIPS$\downarrow$ \\ L1Depth$\downarrow$\end{tabular} 
 &\begin{tabular}{c}13.94\\0.068 \\ 0.506 \\ 2.06 \end{tabular} &
 \begin{tabular}{c}16.98 \\0.132 \\ 0.410 \\ 0.297 \end{tabular} &
 \begin{tabular}{c}15.14 \\0.094 \\ 0.52 \\ 0.588 \end{tabular} &
 \begin{tabular}{c}17.06 \\0.137 \\ 0.394 \\ 0.373 \end{tabular} &
 \begin{tabular}{c}19.76 \\0.215 \\ 0.350 \\ 0.272 \end{tabular} &
    \begin{tabular}{c}16.96 \\ 0.157 \\ 0.408 \\ 0.341\end{tabular} &
    \begin{tabular}{c}13.33 \\0.067  \\ 0.5646 \\ 0.773 \end{tabular} &
    \begin{tabular}{c}15.85 \\0.114 \\ 0.465 \\ 0.690 \end{tabular}

    \\
    \hline    

    \multicolumn{2}{|c|}{NICE-SLAM~\cite{Zhu2022CVPR}}
    &\begin{tabular}{c} PSNR$\uparrow$\\ SSIM$\uparrow$ \\ LPIPS$\downarrow$ \\ L1Depth$\downarrow$\end{tabular} 
 &\begin{tabular}{c}Fail\end{tabular} &\begin{tabular}{c}15.45 \\ 0.573 \\ 0.502 \\ 0.205 \end{tabular} 
 &\begin{tabular}{c}13.86\\0.502 \\ 0.511\\ 0.397\end{tabular} 
 &\begin{tabular}{c}14.25\\0.585 \\0.491 \\0.292 \end{tabular} 
 &\begin{tabular}{c}13.74\\0.610 \\ 0.497\\ 0.384 \end{tabular}

    &\begin{tabular}{c}16.49\\0.634 \\ 0.425\\ 0.191\end{tabular}
    &\begin{tabular}{c}14.72\\0.579 \\ 0.459\\ \bf0.226 \end{tabular}
    &\begin{tabular}{c}14.75\\ 0.580 \\ 0.480 \\ 0.282\end{tabular}

    \\
    \hline   

        \multicolumn{2}{|c|}{iMAP*~\cite{SucaretalICCV2021}~\cite{Zhu2022CVPR}}
    &\begin{tabular}{c} PSNR$\uparrow$\\ SSIM$\uparrow$ \\ LPIPS$\downarrow$ \\ L1Depth$\downarrow$\end{tabular} 
 &\begin{tabular}{c} Fail \end{tabular} 
 &\begin{tabular}{c}13.10\\0.510 \\0.540 \\ 0.269\end{tabular} 
 &\begin{tabular}{c}13.45\\0.465 \\ 0.529\\ 0.391 \end{tabular} 
 &\begin{tabular}{c}13.44\\0.559 \\0.516 \\0.294\end{tabular} 
 &\begin{tabular}{c}11.84\\0.535 \\ 0.563\\0.470\end{tabular}
 
    &\begin{tabular}{c}16.35\\0.609 \\ 0.448\\\bf0.175\end{tabular}
    
    &\begin{tabular}{c}11.53\\0.484 \\ 0.554\\0.586 \end{tabular}
    
     &\begin{tabular}{c}13.28 \\0.527 \\ 0.524 \\0.295\end{tabular}

    \\
    \hline   

    \end{tabular}

    \caption{\textbf{Baseline Comparison.} For each scene, each metric is averaged over the frames.} 
    \label{table:comparison}

\end{table*}

\subsection{Ablation}
\paragraph{Color Only Supervision}
\setlength{\tabcolsep}{1.0mm}
\begin{table}[h]
\centering
    \resizebox{\linewidth}{!}{
    \begin{tabular}{|c|cc|cc|cc|cc|}

    \hline
      \multicolumn{1}{|c|}{}& \multicolumn{2}{c|}{PSNR$\uparrow$} & \multicolumn{2}{c|}{SSIM$\uparrow$} & \multicolumn{2}{c|}{LPIPS$\downarrow$} & \multicolumn{2}{c|}{L1Depth$\downarrow$} 
      \\
    \multicolumn{1}{|c|}{}& Mean & Std & Mean & Std & Mean & Std & Mean & Std    \\
    \hline\hline
    \multicolumn{1}{|c|}{Ours }&
    21.96 & 0.777  & 0.701 &0.032 & 0.167 & 0.016 & 0.233 & 0.026
    \\
    \hline

    \multicolumn{1}{|c|}{Ours w/o depth}&
    22.55 & 0.854 & 0.725 &0.033 &0.148 &0.013 &0.726 & 0.169 
    \\
    \hline

    \end{tabular}
    }
    \caption{\textbf{Depth Input Ablation for On-the-Fly Mapping.}} 
\label{table:mono}

\end{table}

Table~\ref{table:mono} shows ablation analysis of the on-the-fly performance result without depth supervision averaged over all sequences from Table~\ref{table:onthefly}. While we observe a drop in geometry prediction quality, view synthesis results are very similar. This demonstrates that our method can also be used for photorealistic mapping by using large-scale Monocular SLAM system with RGB streams as input. 

\paragraph{Feature Propagation}
In Table~\ref{table:feature_propagation} we ablate our feature propagation strategy (see~Sec.~\ref{subsubsec:modelcreation}). It improves all metrics for both, RGB and RGBD inputs, and indicates the importance of propagating the trained information between models for better initialization.

\setlength{\tabcolsep}{0.55mm}
\begin{table}[h]
\centering
    \resizebox{\linewidth}{!}{
    \begin{tabular}{|c|cc|cc|cc|cc|}

    \hline
      \multicolumn{1}{|c|}{}& \multicolumn{2}{c|}{PSNR$\uparrow$} & \multicolumn{2}{c|}{SSIM$\uparrow$} & \multicolumn{2}{c|}{LPIPS$\downarrow$} & \multicolumn{2}{c|}{L1Depth$\downarrow$} 
      \\
    \multicolumn{1}{|c|}{}& Mean & Std & Mean & Std & Mean & Std & Mean & Std    \\
    \hline\hline
    
    \multicolumn{1}{|c|}{RGBD \begin{tabular}{c} w/o propagation \\ Ours\end{tabular}  }
    &\begin{tabular}{c}21.84 \\ \bf21.96 \end{tabular} &\begin{tabular}{c}0.799 \\ \bf0.777\end{tabular} 
    
    &\begin{tabular}{c} 0.692\\ \bf0.701\end{tabular} &\begin{tabular}{c} 0.034\\\bf0.032\end{tabular}
    
    &\begin{tabular}{c}0.175 \\ \bf0.167\end{tabular}
    &\begin{tabular}{c} 0.017\\ \bf0.016\end{tabular} 
    
    &\begin{tabular}{c}0.239 \\ \bf0.233\end{tabular} &\begin{tabular}{c}0.033 \\ \bf0.026\end{tabular} 
    \\
    \hline
    \hline

    \multicolumn{1}{|c|}{RGB \begin{tabular}{c} w/o propagation \\ Ours\end{tabular}  }
    &\begin{tabular}{c} 22.40 \\ \bf22.55 \end{tabular} 
    &\begin{tabular}{c}0.885 \\\bf0.854 \end{tabular}
   
    &\begin{tabular}{c}0.721\\\bf0.725 \end{tabular} &\begin{tabular}{c}0.034\\\bf0.033\end{tabular}

    &\begin{tabular}{c}0.151\\\bf0.148 \end{tabular} 
    &\begin{tabular}{c} \bf0.013 \\\bf0.013 \end{tabular} 

    &\begin{tabular}{c} 0.790 \\\bf0.726 \end{tabular} &\begin{tabular}{c}0.217\\ \bf0.169\end{tabular} 
    
    \\
    \hline

    \hline    

    \end{tabular}
    }
    \caption{\textbf{Ablation of our Feature Propagation.}}
    \label{table:feature_propagation}

\end{table}

\paragraph{Memory Consumption}
We report memory size used for the map representation in Table~\ref{table:memory_consumption} and the performance with different map sizes in Table~\ref{table:memory_ablation}. We use up to 12 models for the map representation and therefore the maximum map size is 12 times the NeRF model. The map size ablation in Table~\ref{table:memory_ablation} shows our method still shows competitive performance even when we limit the map size thanks to the flexible loop closure handing and effective feature allocation by our proposed spherical encoding.

\section{Conclusion}
We have proposed NEWTON, a novel neural-field based mapping method which can handle online and dynamic state updates from the camera tracking module. We efficiently allocate and train multiple local neural fields with the propsoed parametarization and model training strategy. The strong advantage of our method is its robustness and flexibility to large pose updates by loop closure and its superior performance on real-world large-scale scenes compared to existing world-centric approaches. As future work, we will investigate a loop-closable end-to-end neural field SLAM system that includes camera localization by using the proposed scene representation.

\paragraph{Limitations}
We use a radiance field-based representation which is not ideal for geometry reconstruction. 
To improve the usefulness of the mapping for camera pose tracking, we will investigate the use of more surface-oriented neural field representations like SDFs or occupancy fields. 
Further, the time series plot in Fig.~\ref{fig:timeseries} indicates that, while our approach is significantly less affected by loop closure than the baseline model, we still observe a small drop in performance because ORB-SLAM's pose graph optimization distributes the drift correction to all keyframes and relative pose updates in the models still happen.
However, this can be mitigated by exchanging the loop closure module with \eg hierarchical pose graph optimization where large drift correction by loop closure is done only for the selected model-center keyframe and the camera poses of local training frames are separately optimized similar to~\cite{dai2017bundlefusion}.

\begin{table}[t]
\centering
\begin{tabular}{|l|l|}
\hline
Method & Memory Size (MB) \\
\hline
\hline
\begin{tabular}{l}iMAP~\cite{SucaretalICCV2021} \end{tabular} & 1.02 \\
\hline
\begin{tabular}{l}NICE-SLAM ~\cite{Zhu2022CVPR} \end{tabular} & 12.0 (16cm) \\
\hline
\begin{tabular}{l}Unbounded Instant-NGP ~\cite{nerfstudio} \end{tabular} & 67.1 \\
\hline
\begin{tabular}{l}Ours (up to 12 models) \end{tabular}& 805.2
\\ 
\hline
\end{tabular}
\caption{\textbf{Mapping Representation Memory Usages.}} 
\label{table:memory_consumption}
\end{table}

\setlength{\tabcolsep}{0.55mm}
\begin{table}[t]
\centering
     \resizebox{\linewidth}{!}{
    \begin{tabular}{|cc|cc|cc|cc|cc|}

    \hline
      \multicolumn{2}{|c|}{}& \multicolumn{2}{c|}{PSNR$\uparrow$} & \multicolumn{2}{c|}{SSIM$\uparrow$} & \multicolumn{2}{c|}{LPIPS$\downarrow$} & \multicolumn{2}{c|}{L1Depth$\downarrow$} 
      \\
    \multicolumn{2}{|c|}{}& Mean & Std & Mean & Std & Mean & Std & Mean & Std    \\
    \hline\hline
    
    \begin{tabular}{c} World-Centric \\ Single Model\end{tabular}  & \begin{tabular}{c} 67MB \\ 805MB\end{tabular}  
    &\begin{tabular}{c} 21.46 \\ 21.57 \end{tabular} &\begin{tabular}{c}1.67 \\ 1.77\end{tabular} 
    
    &\begin{tabular}{c} 0.690\\ 0.692\end{tabular} &\begin{tabular}{c} 0.069\\0.071\end{tabular}
    
    &\begin{tabular}{c}0.162 \\ \bf0.158\end{tabular}
    &\begin{tabular}{c} 0.023\\ 0.023\end{tabular} 
    
    &\begin{tabular}{c}0.238 \\ 0.240 \end{tabular} &\begin{tabular}{c}0.049 \\ 0.051\end{tabular} 
    \\
    \hline
    Ours & \begin{tabular}{c} 67MB \\ 805MB\end{tabular}  
    &\begin{tabular}{c}21.75 \\ \bf21.96 \end{tabular} &\begin{tabular}{c}0.841 \\ \bf0.777\end{tabular} 
    
    &\begin{tabular}{c} 0.687\\ \bf0.701\end{tabular} &\begin{tabular}{c} \bf0.031\\0.032\end{tabular}
    
    &\begin{tabular}{c}0.180 \\ 0.167\end{tabular}
    &\begin{tabular}{c} \bf0.016\\ \bf0.016\end{tabular} 
    
    &\begin{tabular}{c}0.235 \\ \bf0.233\end{tabular} &\begin{tabular}{c}\bf0.026 \\ \bf0.026\end{tabular} 
    
    \\
    \hline

    \hline    

    \end{tabular}
    }
    \caption{\textbf{Different Map Sizes for On-the-Fly Mapping.}} 
\label{table:memory_ablation}
\end{table}

{\small
\bibliographystyle{ieee_fullname}
\bibliography{bib_long,bibliography,egbib}
}

\end{document}